\documentclass{article}

\usepackage[final, nonatbib]{tccml_neurips_2020}

\usepackage[utf8]{inputenc} 
\usepackage[T1]{fontenc}    
\usepackage{hyperref}       
\usepackage{url}            
\usepackage{booktabs}       
\usepackage{amsfonts}       
\usepackage{nicefrac}       
\usepackage{microtype}      
\usepackage{graphicx}
\usepackage{tikz}
\usepackage{svg}
\usepackage{glossaries}
\usepackage{wrapfig}
\usepackage{comment}
\usetikzlibrary{positioning}
\hypersetup{
    colorlinks=true,
    linkcolor=blue,
    filecolor=magenta,      
    urlcolor=cyan,
}

\title{Machine Learning for Glacier Monitoring in the Hindu Kush Himalaya}

\usepackage{authblk}
\author[1, *]{Shimaa Baraka}
\author[1, *]{Benjamin Akera}
\author[2, *]{Bibek Aryal}
\author[3]{Tenzing Sherpa}
\author[3]{Finu Shresta}
\author[4]{Anthony Ortiz}
\author[5]{Kris Sankaran}
\author[4]{Juan Lavista Ferres}
\author[3]{Mir Matin}
\author[1]{Yoshua Bengio}
\affil[1]{Mila}
\affil[2]{University of Texas at El Paso}
\affil[3]{Integrated Centre for Mountain Development}
\affil[4]{Microsoft AI for Good Research Lab}
\affil[5]{University of Wisconsin - Madison}
\affil[*]{Equal Contribution}

\begin{document}
\newacronym{hkh}{HKH}{Hindu Kush Himalayan}
\newacronym{ndsi}{NDSI}{Normalised Difference Snow Index}
\newacronym{ndvi}{NDVI}{Normalised Difference Vegetation Index}
\newacronym{ndwi}{NDWI}{Normalised Difference Water Index}
\newacronym{icimod}{ICIMOD}{International Centre for Integrated Mountain Development}
\maketitle

\begin{abstract}
Glacier mapping is key to ecological monitoring in the \gls{hkh} region. Climate change poses a risk to individuals whose livelihoods depend on the health of glacier ecosystems. In this work, we present a machine learning based approach to support ecological monitoring, with a focus on glaciers. Our approach is based on semi-automated mapping from satellite images. We utilize readily available remote sensing data to create a model to identify and outline both clean ice and debris-covered glaciers from satellite imagery. We also release data and develop a web tool that allows experts to visualize and correct model predictions, with the ultimate aim of accelerating the glacier mapping process.

\end{abstract}

\section{Introduction}

Glaciers are a source of freshwater and are critical to the Hindu Kush Himalayan (HKH) region both ecologically and societally \cite{bajracharya2006impact}. However, glaciers are continuing to shrink at an alarming rate and this will result in diminished freshwater flow. This is likely to cause adverse effects for the benefactors of freshwater flow from glaciers.
Additionally, glacier shrinkage has been shown to be a significant factor in the current sea-level rise \cite{dyurgerov2005mountain}.This calls for effective and efficient methods to map and delineate glaciers in order to monitor changes and plan integrated water resource management and glacial hazard and risk management. 
 
In such areas, remote sensing offers complementary information that can be used to monitor glaciers \cite{racoviteanu2009challenges, paul2013accuracy}. Remote sensing allows the estimation of parameters like snow cover, glacier elevation, and ice index over large geographical and temporal scales. 
Utilising this information, different automated methods of delineating glaciers have been developed. However, the efficacy and accuracy of these methods are affected by cloud cover, highly variable snow conditions, and the spectral similarity of supra-glacial debris with moraines and bedrock~\cite{biddle2015mapping}. These errors are somewhat addressed through the application of semi-automated mapping methodologies, which combine outputs from automated methods with manual interventions. However, this is labor intensive and time-consuming. Machine learning techniques can play a significant and positive role in speeding the process up. 

We apply machine learning techniques to automate methods for glacier mapping from satellite imagery. We utilize semantic segmentation - a deep learning approach that performs pixel-wise classification in images.
Using the \gls{hkh} glacier region as an area of study, we use available satellite imagery from Landsat  and glacier labels provided by the \gls{icimod} - a regional intergovernmental learning and knowledge sharing center serving the eight regional member countries of the \gls{hkh}~\cite{bajracharya2011status}. Using these resources, we develop an extensible pipeline, a dataset, and baseline methods that can be utilized for automated glacier mapping from satellite images. We also present qualitative and quantitative results describing properties of our models. Additionally, we deploy our models as a web-based tool to demonstrate how machine learning can complement, rather than supplant, existing workflows.

\section{Study Area and Data Sources}

Our experiments are carried out on the \gls{hkh} region. The \gls{hkh} is also known as the Third Pole of the world as it consists of one of the largest concentrations of snow and ice besides the two poles. It constitutes more than 4 million square kilometers of hills and mountains in the eight countries of Afghanistan, Bangladesh, Bhutan, China, India, Myanmar, Nepal and Pakistan. Glaciers have been periodically identified and classified by experts at the \gls{icimod}~\cite{bajracharya2011status}. Glaciers are categorized as either \textit{clean ice} or \textit{debris-covered} subtypes.

The labels we use have been generated through a semi-automated pipeline based on hyperpixel segmentation. Historically, \gls{icimod} has used the eCognition software~\cite{ecognition2020} to segment  Landsat imagery into image objects defined by a contiguous set of pixels with similar intensity value. Those hyperpixels that contain debris or ice glacier are then merged and downloaded for refinement. The manual refinement phase involves removing labeled regions that are not at plausible glacier elevations or which do not pass specified \gls{ndvi}, \gls{ndsi} or \gls{ndwi} thresholds~\cite{goward1991normalized,hall2010normalized,gao1996ndwi}.


We release our data in the \href{http://lila.science/datasets/hkh-glacier-mapping}{LILA BC repository}. The input data come in two forms -- the original 35 Landsat tiles and 14,190 extracted numpy patches. Labels are available as raw vector data in shapefile format and as multichannel numpy masks. Both the labels and the masks are cropped according to the borders of \gls{hkh}. The numpy patches are all of size $512 \times 512$ and their geolocation information, time stamps, and source Landsat IDs are available in a geojson metadata file. 

\section{Model Architecture and Methodological Pipeline}

The task of identifying and mapping glaciers in remote sensing images fits well within the framework of semantic segmentation. We adapted the U-Net architecture for this task~\cite{ronneberger2015u}. The U-Net is a fully convolutional deep neural network architecture; it consists of two main parts, an encoder network and a decoder network. 
The encoder is a contracting path that extracts features of different levels through a sequence of downsampling layers making it possible to capture the context of each pixel while the decoder is an expanding sequence of upsampling layers that extracts the learned encoded features and upsamples them to the original input resolution. Skip connections are employed between the corresponding encoder and decoder layers of the network to enable efficient learning of features by the model without losing higher resolution spatial information because of low spatial resolution in the bottleneck between encoder and decoder. 

The model was trained using gradient descent and the Dice loss~\cite{sudre2017generalised} was used as the optimization criterion (see the Appendix). We adapt a human-in-the-loop approach to correct the segmentation errors made by the model. This is useful because glacier mapping often requires expert opinion and models make errors that need to be resolved by people. 

\begin{figure}[ht]
\centering
     \includegraphics[width=1.1\textwidth]{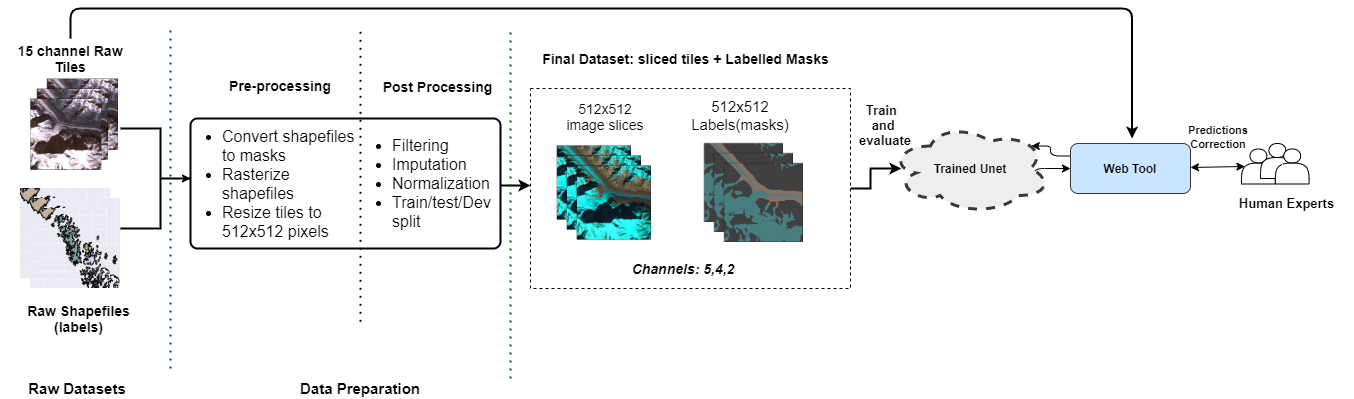}
      \caption{Our methodological pipeline first converts LE7 tiles to preprocessed patches used for model training. The trained model is deployed as an interactive web tool.}
      \label{methodology}
\end{figure}

Our approach is summarized in a multi-step pipeline presented in Figure \ref{methodology}. It first converts the raw tiles into patches and converts their vector data labels to masks. We filter, impute and normalize the resulting patch-mask pairs before splitting them into train, test and validation data sets. The code to replicate our process can be found in a GitHub repository\footnote{\href{http://github.com/krisrs1128/glacier_mapping/}{https://github.com/krisrs1128/glacier\_mapping/}}. The script to query Landsat 7 tiles using Google Earth engine is in another GitHub repository\footnote{\href{https://github.com/Aryal007/GEE_landsat_7_query_tiles/}{https://github.com/Aryal007/GEE\_landsat\_7\_query\_tiles/}}.

\section{Experiments}
\label{sec:experiments}

In this section, we characterize the performance of existing methods on tasks related to glacier segmentation. We intend to provide practical heuristics and isolate issues in need of further study.

\paragraph{Band Selection}

Model performance tends to deteriorate in the many-bands limited-training-data regime~\cite{ortiz2018integrated}. This is often alleviated through band subset selection. Here, we study whether specific channels are more relevant for glacier mapping. We experimented with the combination of bands B5 (Shortwave infrared), B4 (Near infrared), and B2 (Green) which is the false-color composite combination used to differentiate snow and ice from the surrounding terrain when manually delineating glaciers. We compare this with (1) the true color composite band combination, B1 (Blue), B2 (Green), B3 (Red) and (2) all Landsat 7 bands. We also consider (1) slope and elevation from the Shuttle Radar Topography Mission (SRTM) as additional channels and (2) spectral indices - snow index (\gls{ndsi}), water index (\gls{ndwi}), and vegetation index (\gls{ndvi}) - as used in manual glacier delineation~\cite{bajracharya2011status}. Lastly, we perform pixel-wise classification on all channels with a random forest (RF) and select channels with feature importance scores greater than 5\%, see appendix Figure \ref{fig:feat_importance_rf}.

\begin{figure}
    \centering
    \includegraphics[width=0.85\textwidth]{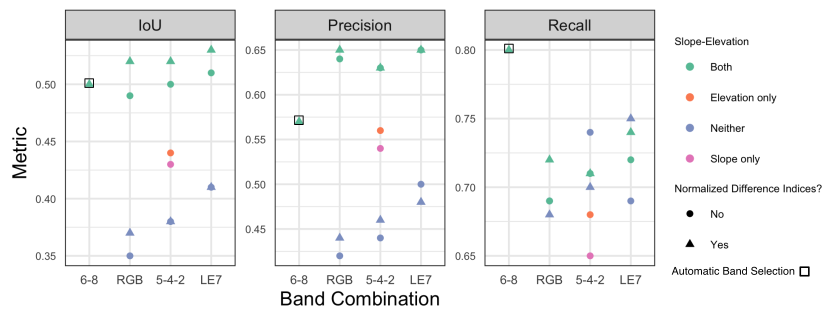}
    \caption{Experimental results for channel selection. The $x$-axis indicates which LE7 bands were used. Color describes whether elevation and slope were used. Runs using \gls{ndwi}, \gls{ndsi}, and \gls{ndvi} are labeled with triangles. Elevation and slope data significantly boost performance, and using all bands is better than using any subset. Results when using RF features are enclosed in a square.}
    \label{fig:channel_selection_results}
\end{figure}

\begin{figure}[ht]
\centering
     \includegraphics[width=0.7\textwidth]{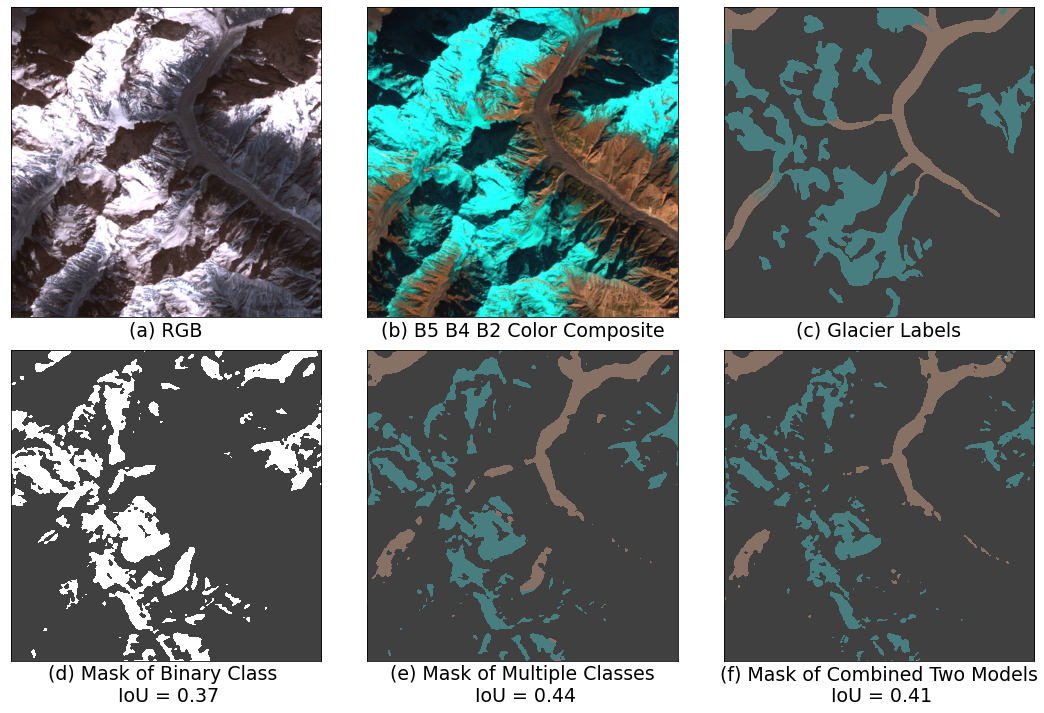}
      \caption{Example imagery and labels (a - c), with model predictions (d - f). Blue and orange labels are for clean ice and debris-covered glaciers, respectively. (d) A model trained to recognize the union of clean ice or debris-covered   glaciers fails to recognize major debris-covered glaciers. (e - f) Multiclass and combined binary models give comparable predictions.}
      \label{fig:debris_clean}
\end{figure}

Figure~\ref{fig:channel_selection_results} shows performance when varying input channels.
The experiments are carried out on the 383 patches with at least 10\% of pixels belonging to either clean ice or debris-covered glaciers. We evaluated the model over 55 patches using Intersection over Union (IoU). The RF classifier features did not achieve the maximum IoU, likely due to a lack of spatial context. Adding elevation and slope channels provides an improvement of 10-14\% IoU. This agrees with domain knowledge -- elevation and slope maps are referred to in the current process. Appendix Figure \ref{fig:glacier_elev} illustrates that the model learns that low elevation and steep areas typically do not contain glaciers. Using \gls{ndvi}, \gls{ndsi}, and \gls{ndwi} improves results when input channels are different from those used to define the indices.

\paragraph{Debris covered versus clean ice glaciers}
There are two types of glaciers we care about: clean ice glaciers and debris-covered glaciers. Clean ice glaciers have an appearance similar to snow. Debris-covered glaciers are covered in a layer of rock and flow through valley-like structures. For segmentation, clean ice glaciers are often confused with snow, resulting in false positives. Debris-covered glaciers are more similar to the background, often leading to false negatives. Debris-covered glaciers are also much rarer. We experimented with binary and multiclass approaches to segmentation.

We trained a 2-class model to segment glacier from background areas and compared it with 3-class model for clean ice vs. debris-covered vs. background. We also compared the 3-class model with two binary models for each glacier type. We filtered to patches where both debris-covered and clean ice glaciers were present, resulting in 648 training patches and 93 validation patches. Since many patches contain few positive class pixels, we evaluate IoU over the whole validation set rather than the mean IoU per patch. Table \ref{table:debris_vs_ice} shows that the multiclass model and binary model deliver comparable overall performance.  However, the approaches differ in regions with higher coverage from debris-covered glaciers. Table \ref{table:debris_per} and figure \ref{fig:debris_clean} show an increase in the performance gap in favour of the multiclass model as the debris-covered glacier percentage increases.

\section{Glacier Mapping Tool}

 \begin{figure*}[!tbp]
        \centering 
        \includegraphics[width=0.4\textwidth]{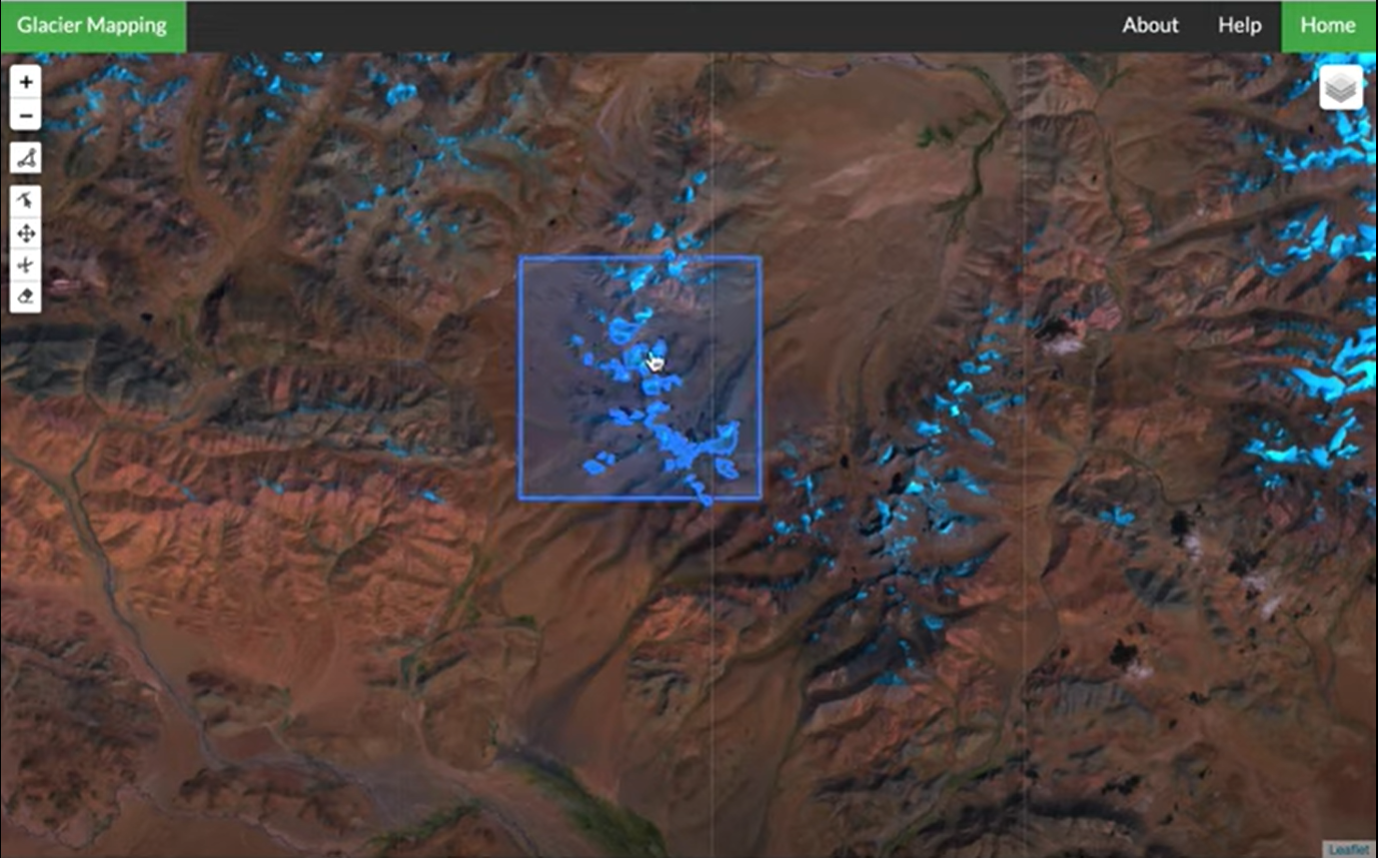}
        \includegraphics[width=0.4\textwidth]{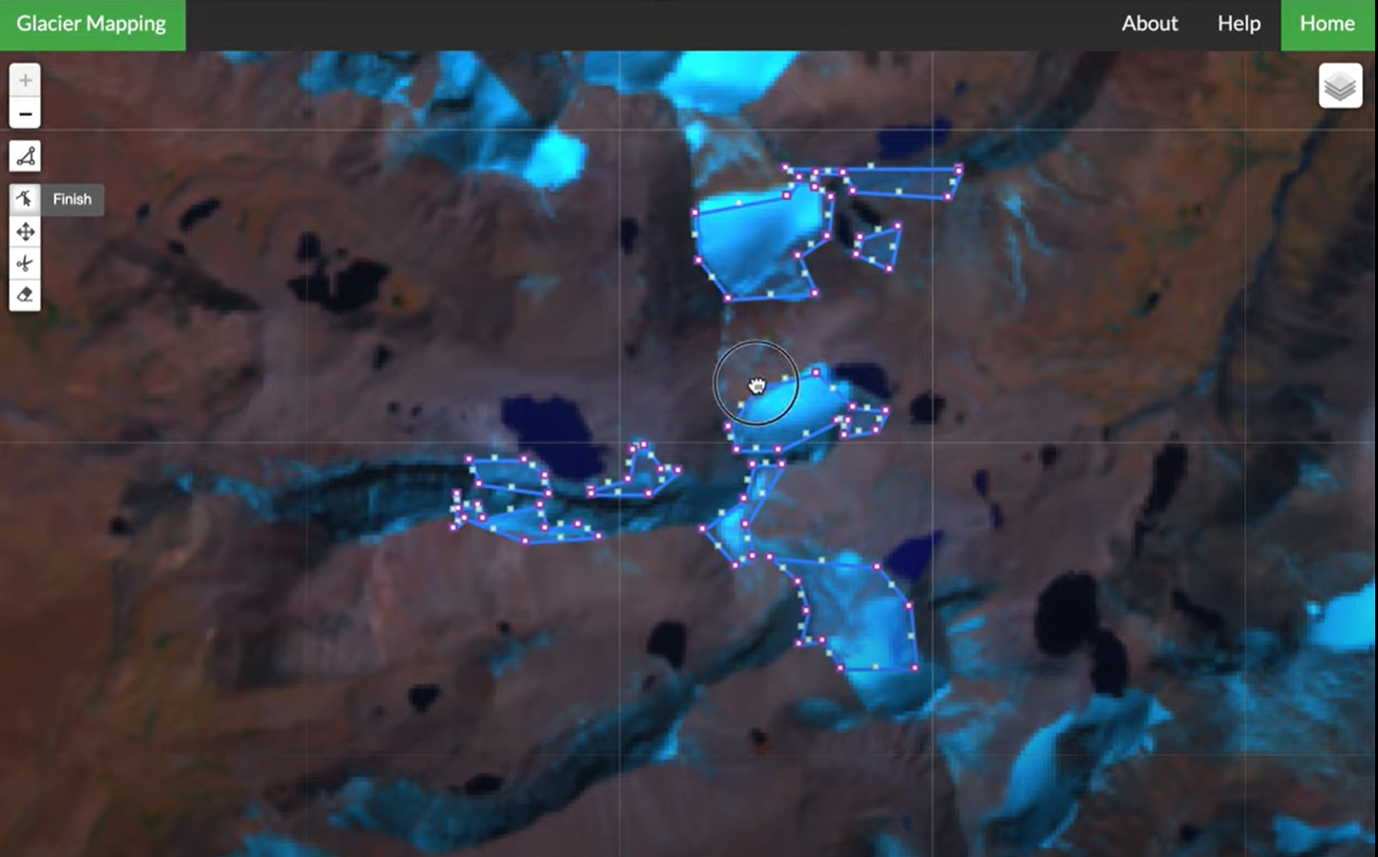}
		\caption{The image on the left shows the polygonized prediction for an area of interest. The image to the right shows the tool's functionality of allowing users to correct predictions.}
		\label{fig:tool_screenshot}
\end{figure*}

To support the work of geospatial information specialists to delineate glaciers accurately we developed an interactive glacier mapping tool. The tool allows users to test our segmentation models on different sources of satellite imagery. Users can visualize predictions in the form of polygons and edit them to obtain a glacier map for the area of interest. This interactivity supports the process of validating models, identifying systematic sources of error, and refining predictions before release. Users can compare data sources, which can clarify ambiguities. As future work, we intend to incorporate model retraining functionality. A screenshot from the tool is visible in Figure \ref{fig:tool_screenshot}.

\section{Conclusion and Future Work}

We have presented deep learning and remote sensing techniques that support semi-automated glacier mapping. We have experimentally explored the effects of channel selection and task definition on performance.
Finally, we describe a web tool to provide feedback and correct errors made by the model. More work needs to be done to (1) incorporate the human feedback into the trained model through some form of active learning, (2) develop network architectures and criteria that better use domain knowledge, and (3) understand the generalizability of these methods to regions outside of the Hindu Kush Himalaya.

\section{Acknowledgements}
We acknowledge \gls{icimod} for providing a rich dataset which this work has been built on. We also appreciate Microsoft for funding this project under the AI for Earth program. This research was enabled in part by support provided by Calcul Quebec and Compute Canada. We would like to thank Dan Morris from Microsoft AI for Earth for making this collaboration between ICIMOD and academia possible
\bibliographystyle{plain}
\bibliography{citations}

\clearpage

\appendix

\section{Implementation Details}

We query Landsat 7 raw images used for creating labels~\cite{bajracharya2011status} using Google Earth Engine. In addition to the raw Landsat 7 tiles, we compute Normalized-Difference Snow Index (NDSI), Normalized-Difference Water Index (NDWI), and Normalized-Difference Vegetation Index (NDVI) and add them as additional bands to the tiles. Finally, we query slope and elevation from the Shuttle Radar Topography Mission~\cite{doi:10.1029/2005RG000183} and add them as additional bands to give us final tiles with 15 bands. The vector data corresponding to glacier labels~\cite{https://doi.org/10.26066/rds.31029} is downloaded from \gls{icimod} \href{https://rds.icimod.org/}{Regional Database System} (RDS). We then follow pre-processing and post-processing as shown in Figure~\ref{methodology} to prepare the data. The pre-processing steps include conversion of vector data to image masks, cropping the input image and vector data to HKH borders, and slicing the mask and tiles to patches of size $512\times 512$ pixels. We then filter patches with low glacier density (thresholds vary by experiment), impute nan values with 0, normalize across channel for each patch, and randomly split the data into train (70\%) / dev (10\%) / test (10\%).

We make use of a U-Net architecture~\cite{ronneberger2015u} for the segmentation of glacier labels. We use a kernel size of $3\times 3$ for convolution layers in the downsampling operations and kernel size of $2\times 2$ for convolution layers and transpose convolution layers in the upsampling layers. For the pooling operation, we use maxpool with kernel size $2\times 2$. The output of the first convolution operation has 16 channels and we double the channels after each convolutional layer in during downsampling and in the bottleneck layer. We halve the output channels in each convolutional layer during upsampling. We use a depth of 5 meaning there are 5 downsampling layers followed by 5 upsampling layers with a bottleneck layer in between. We use Adam as our optimizer with a learning rate of $1e^{-4}$. We use spatial dropout~\cite{tompson2015efficient} of 0.3 and $\ell^1$ regularization with $\lambda = 5e^{-4}$ to prevent the model from overfitting on training data. 

\section{Supplemental Tables and Figures}

\begin{figure}[ht]
\centering
     \includegraphics[width=0.7\textwidth]{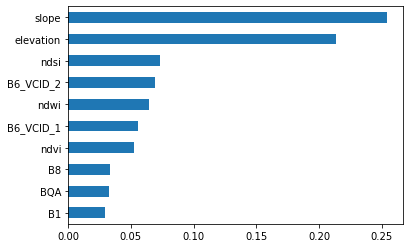}
      \caption{Feature importance scores using Random Forest. Slope and elevation are key variables.}
       \label{fig:feat_importance_rf}
\end{figure}

\begin{table}[!h]
\caption{A comparison of IoU for the U-Net based model architecture with other traditional machine learning approaches. Pixels from train slices are sampled to train machine learning classifiers to predict labels for each pixel. There are a total of 15 features for each pixels, one for value in each band, and one of the three output labels. The output segmentation mask is generated by predicting the class for each pixel in the test slices using trained classifiers. The U-Net based classifier outperforms conventional machine learning classifiers especially in case of debris glaciers.}
\label{table:comparison_between_models}
\centering
\begin{tabular}{|l|c|c|c|}
\hline
\textbf{Model} & \textbf{IoU of Clean Ice Glaciers} & \textbf{IoU of Debris Glaciers} \\
\hline
Random Forest & 0.5807 & 0.2024 \\
\hline
Gradient Boosting & 0.5663 & 0.1930 \\
\hline
Multi-Layered perceptrons & 0.5452 & 0.1781 \\
\hline
U-Net & 0.5829 & 0.3707 \\
\hline
\end{tabular}
\end{table}

\begin{figure}[ht]
\centering
     \includegraphics[width=1.0\textwidth]{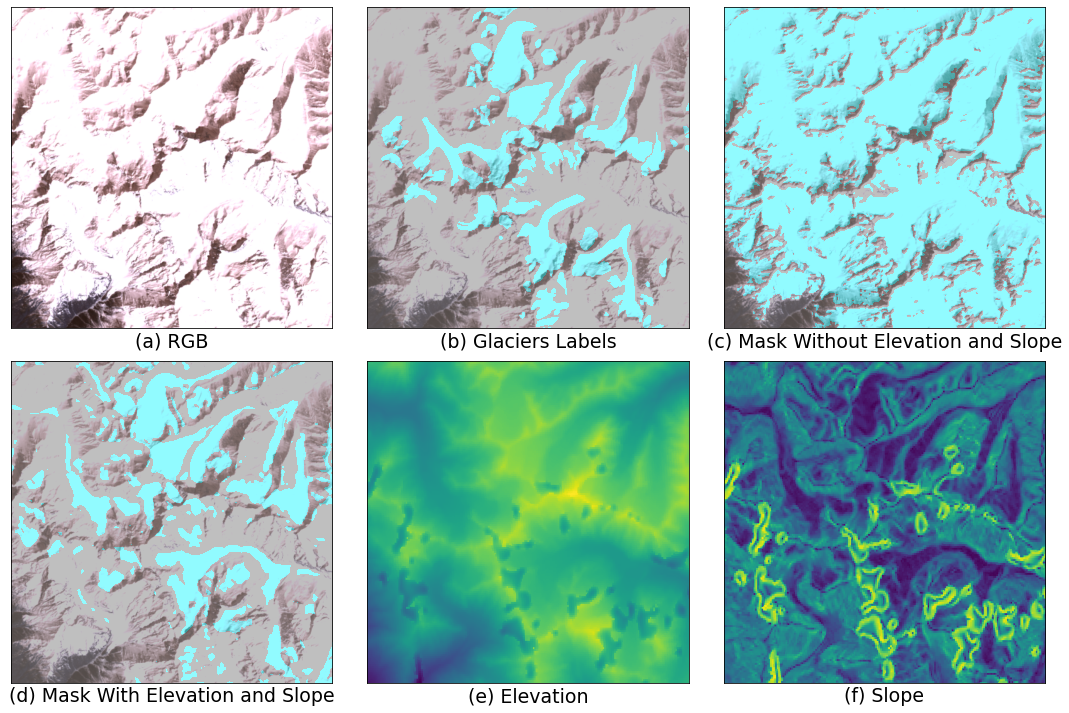}
      \caption{Slope and elevation are critical for improving precision of glacier predictions. Light blue labels represent both clean ice and debris-covered glaciers. Light yellow and green are areas of high elevation (e) and slope (f). Clean ice glaciers tend to be found at high elevation, while debris-covered glaciers are found in valleys. Neither type of glacier is found on steep slopes.}
       \label{fig:glacier_elev}
\end{figure}

\begin{table}[!h]
\caption{A comparison of error rates on clean ice and debris-covered glaciers across three modeling approaches. The first row is a
model trained to predict glacier or background, without distinguishing between debris-covered or ice glaciers. The second row is a 
multiclass model trained to simultaneously segment debris-covered and clean ice glacier. The final row gives the result of training 
two separate models to distinguish each type of glacier. Results are comparable across approaches, with a slight edge for the split
training approach.}
\label{table:debris_vs_ice}
\centering
\begin{tabular}{|l|c|c|c|}
\hline
\textbf{Model} & \textbf{IoU of Glaciers} & \textbf{IoU of Clean Ice Glaciers} & \textbf{IoU of Debris Glaciers} \\
\hline
Binary Model  & 0.476 & - & - \\
\hline
Multiclass Model & 0.473 & 0.456 & 0.291 \\
\hline
Two Binary Models & 0.48 & 0.476 & 0.31 \\
\hline
\end{tabular}
\end{table}

\begin{figure}
\centering
    \includegraphics[width=0.7\textwidth]{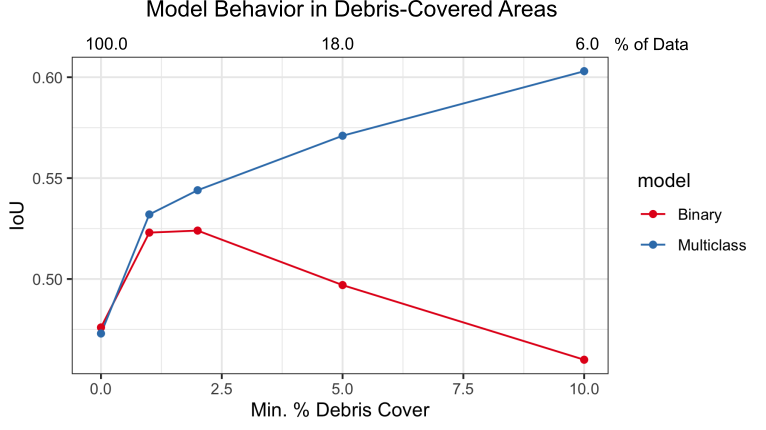}
\caption{Effect of debris percentage on IoU. The multiclass model performs well on areas with high density of debris-covered glaciers. 
The binary model trained to distinguish any type of glacier from background suffers in these regions. When making no distinguish between
glacier types, the model only learns to recognize clean-ice glaciers.}
\label{fig:debris_iou}
\end{figure}

\begin{table}[!h]
\caption{The exact numbers used in Figure \ref{fig:debris_iou}}.
\label{table:debris_per}
\centering
\begin{tabular}{|c|c|c|c|c|}
\hline
\textbf{\%  of Debris} & \textbf{\%  of Data} & \textbf{Binary Class IoU} & \textbf{Multiclass IoU} & \textbf{IoU Difference}\\
\hline
> 0\% & 100\% & 0.476 & 0.473 & -0.3\%\\
\hline
> 1\% & 77\% & 0.523 & 0.532 & +0.9 \%\\
\hline
> 2\% & 52\% & 0.524 & 0.544 & +2\%\\
\hline
> 5\% & 18\% & 0.497 & 0.571 & +7.4\%\\
\hline
> 10\% & 6\% & 0.46 & 0.603& +14.3\%\\
\hline
\end{tabular}
\end{table}

\begin{figure}
    \centering
    \includegraphics[width=0.8\textwidth]{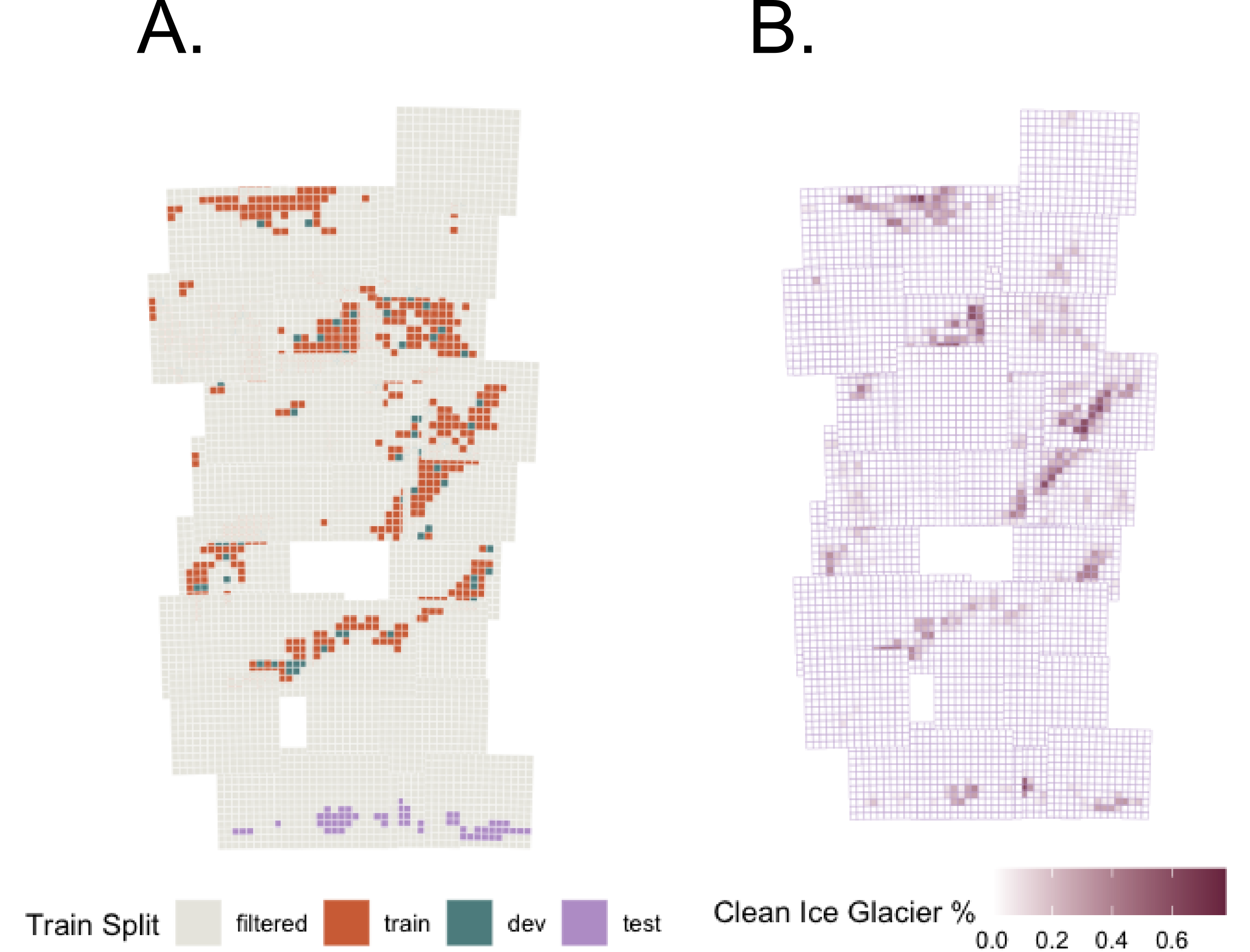}
    \caption{(a) One of three geographically disjoint train and test sets used in a supplemental experiment on geographic generalization (results in Figure \ref{fig:geo_iou}). Each large square is a LE7 tile, each shaded element is a $512\times 512$ patch. Patches without any glaciers are filtered as in prior experiments. The validation set is chosen at random from the same region as training. (b) The proportion of pixels covered by clean ice glacier within each patch. Note that the test region is less densely covered by glaciers.}
    \label{fig:geo_split}
\end{figure}

\begin{figure}
    \centering
    \includegraphics[width=0.7\textwidth]{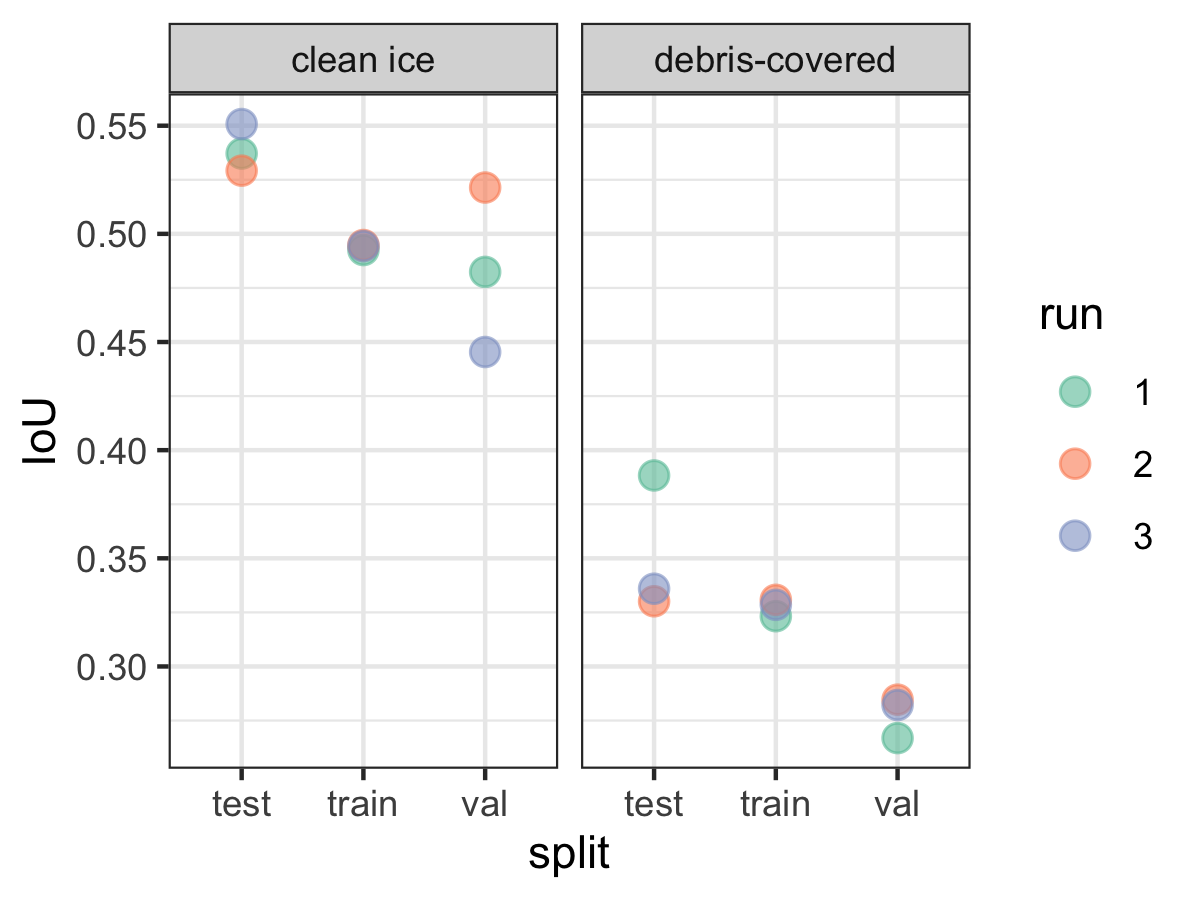}
    \caption{Results of a supplemental geographic generalization experiment using the multiclass model of Section \ref{sec:experiments}. Train and test sets are chosen at random and required to be geographically disjoint, as in Figure \ref{fig:geo_split}a. Surprisingly, performance slightly increases in the test set. Evidently, glacier appearance is relatively homogeneous across the area of study. The increase in performance can be explained by the fact that, in each random geographic split, the test region had a lower glacier density, see Figure \ref{fig:geo_split}b.
}
\label{fig:geo_iou}
\end{figure}

\end{document}